\documentclass[11pt]{article}
\usepackage{eamt20}
\usepackage{times}
\usepackage{url}
\usepackage{latexsym}
\usepackage[small,bf]{caption} 
\usepackage{caption}    
\usepackage{subcaption} 
\usepackage[export]{adjustbox}
\usepackage{multirow}
\usepackage{booktabs}
\usepackage[T1]{fontenc}
\usepackage{bbm}        
\usepackage{amsmath}
\usepackage[title]{appendix}
\usepackage[flushleft]{threeparttable}  
\usepackage{xcolor}

\newcommand{\rtt}{{\scshape RTT}}
\newcommand{\rttbleu}{{\scshape RTT-Bleu}}
\newcommand{\rttchrf}{{\scshape RTT-chrF}}
\newcommand{\rttbert}{{\scshape RTT-Bert}}
\newcommand{\rttsbert}{{\scshape RTT-SBert}}
\newcommand{\rttbertscore}{{\scshape RTT-BERTScore}}
\newcommand{\bleu}{{\scshape Bleu}}
\newcommand{\sentbleu}{{\scshape sentBLEU}}
\newcommand{\chrf}{{\scshape chrF}}
\newcommand{\bert}{{\scshape Bert}}
\newcommand{\bertscore}{{\scshape BERTScore}}
\newcommand{\sbert}{{\scshape SBert}}
\newcommand{\yisi}{{\scshape YiSi}}

\title{Revisiting Round-Trip Translation for Quality Estimation}

\author{Jihyung Moon \\ Naver Papago \\  \And
  Hyunchang Cho \\ Naver Papago \\
  \\\tt{\{jihyung.moon, hyunchang.cho, lucypark\}@navercorp.com} \\ \And
  Eunjeong L. Park \\ Naver Papago}
 
\begin{document}
\maketitle
\begin{abstract}
Quality estimation (QE) is the task of automatically evaluating the quality of translations without human-translated references.
Calculating {\bleu} between the input sentence and round-trip translation (\rtt) was once considered as a metric for QE, however, it was found to be a poor predictor of translation quality.
Recently, various pre-trained language models have made breakthroughs in NLP tasks by providing semantically meaningful word and sentence embeddings.
In this paper, we employ semantic embeddings to {\rtt}-based QE.
Our method achieves the highest correlations with human judgments, compared to previous WMT 2019 quality estimation metric task submissions.
While backward translation models can be a drawback when using {\rtt}, we observe that with semantic-level metrics, {\rtt}-based QE is robust to the choice of the backward translation system.
Additionally, the proposed method shows consistent performance for both SMT and NMT forward translation systems, implying the method does not penalize a certain type of model.
\end{abstract}

\section{Introduction}

\begin{table}[t!]
\begin{adjustbox}{width=\columnwidth,center}
\begin{tabular}{l|l}
\toprule
Input (en)      & \begin{tabular}[c]{@{}l@{}}`We know it won't change\\ students' behaviour instantly.\end{tabular}                   \\ \midrule
Reference (de) & \begin{tabular}[c]{@{}l@{}}Wir wissen, dass es das Verhalten \\ der Studenten nicht sofort ändern wird.\end{tabular} \\ \midrule
Output (de)     & \begin{tabular}[c]{@{}l@{}}\quotedblbase{}Wir wissen, dass es das Verhalten \\ der Schüler nicht sofort ändern wird.\end{tabular} \\ \midrule
Round-trip (en) & \begin{tabular}[c]{@{}l@{}}``We know that it will not change \\ student behavior immediately.\end{tabular}           \\ \midrule
\multicolumn{2}{c}{\begin{tabular}[c]{@{}l@{}}{\scshape RTT-sentBleu}: 14.99 (rank: 1947/1997)\\{\rttsbert}(*): 98.07 (rank: 1001/1997)\\{\rttbertscore}(*): 97.04 (rank: 1033/1997) \end{tabular}} \rule{0pt}{2.6ex}\\   \bottomrule
\end{tabular}
\end{adjustbox}
\caption{\label{tab:example}
A sample of {\rtt}-based evaluation methods with an example from the WMT19 English--German evaluation set. * denotes our proposed semantic-level methods (Detailed definitions are described in Section~\ref{proposed-methods}). Note that {\sentbleu} could not capture the similarity of the input and {\rtt}.
}
\end{table}

A good machine translation (MT) system converts one language to another while preserving the meaning of a sentence. 
Given a pair of well-performing translation systems between two languages, the meaning of a sentence should remain intact even after a round-trip translation (\rtt) -- the process of translating text from the source to target language (forward translation, FT) and translating the result back into the source language (backward translation, BT).
If the MT systems work reasonably well and no human-produced reference translations are provided, using {\rtt} for translation evaluation seems like a natural choice.

However, in the early 2000s, this practice was not recommended to be used as a translation evaluation method~\cite{huang1990machine,somers2005round,van2006unsupervised}. 
This argument was largely supported by the poor correlation between {\bleu}~\cite{papineni2002bleu} for reference and translated output and {\bleu} for input and {\rtt} (We address this method again in Section~\ref{proposed-methods} as {\rttbleu}).
However, {\bleu} only measures surface-level lexical similarity, thus penalizing paraphrased sentences resulting from the round-trip translation as shown in Table~\ref{tab:example}.

On the other hand, human evaluations conducted on input sentences and translated outputs show a significant positive correlation with human evaluations on input sentences and round-trip sentences~\cite{aiken2010efficacy}.
The result implies if a suitable semantic-level metric is provided, {\rtt}-based method can be used for MT evaluation.
Meanwhile, recently introduced pre-trained language models e.g., {\bert}~\cite{devlin2019bert} and {\sbert}~\cite{reimers2019sentence}, are effective for many natural language processing tasks including semantic similarity detection~\cite{cer2017semeval}.
{\bertscore}~\cite{zhang2019bertscore} and {\yisi}~\cite{lo2019yisi} leveraged such models for MT evaluation and confirmed the efficacy. 


In this paper, we revisit {\rtt} with recently proposed semantic-level metrics for MT quality estimation.
Quality estimation (QE) aims to measure how good a translation is without any human-translated references~\cite{fonseca2019findings} as opposed to reference-based metrics such as {\bleu} or {\chrf}. 
Therefore, with these metrics, it is easy to evaluate translations beyond reference-ready domains, e.g., user logs in commercial services. 

We start by investigating {\rtt}-based QE metrics on different BT systems to choose a proper BT system to examine {\rtt}-based methods across different language pairs.
Then we compare the methods on NMT with statistical machine translation (SMT) systems and demonstrate the compatibility of our methods.
Across the experiments, {\rtt}-based QE metrics with semantic-level similarities outperform lexical-based similarity metrics.
We find the results are related to the metric's ability of detecting paraphrases.

The main contributions of this work are as follows:
\begin{itemize}
    \item We reconsider {\rtt} with suitable semantic-level metrics, specifically {\sbert} and {\bertscore} in our settings, and show it can be used to measure translation quality.
    \item We observe {\rtt} methods using {\sbert} and {\bertscore} are robust to the choice of BT systems.
    \item We present {\rtt} with semantic similarity measurements consistently exhibit high-performance across different FT systems: SMT and NMT.
    \item We find the paraphrase detection ability of metrics is related to the performance of {\rtt}-based QE.
\end{itemize}


\section{Related Work}

\subsection{Quality Estimation}

One goal of QE is to estimate the quality for machine translated sentences without reference translations, but the definition of quality has gradually changed.
Traditional QE aimed to estimate the required amount of post-editing efforts for a given translation in the word, sentence or document level. 
In the sentence level, this can be understood as estimating the Human Translation Error Rate (HTER)~\cite{snover2006study}, or the rate of edit operations which include the insertions of words, deletions or replacements.
The recently proposed view of "QE-as-a-metric"~\cite{fonseca2019findings}\footnote{Since WMT20, this was modified to the "sentence-level direct assessment task".} differs from traditional quality estimation in that it directly aims to estimate the absolute score of a translation, and can be directly compared with previous reference-based metrics.
While reference-based metrics easily achieve above 0.9 Pearson correlation with direct human assessments in the system-level and up to 0.4 correlation in the sentence-level, QE-based metrics typically score less~\cite{ma2019results}.

{\yisi}~\cite{lo2019yisi} is the best performing QE metric from the recent QE-as-a-metrics subtask submitted to the quality estimation shared task of WMT19~\cite{ma2019results}\footnote{We excluded UNI and its variants from consideration, since they do not have any open publications to refer to. See Table 2 in \cite{ma2019results}.}.
It takes contextual embeddings extracted from {\bert} and computes F-scores of semantic phrases using the cosine similarity of words weighting by their inverse document frequency (idf).
{\yisi} has variants for both situations where the references exist ({\yisi}-1) or does not exist ({\yisi}-2).

\subsection{Round-trip Translation}

{\rtt} had frequently been used for a means of evaluating MT systems until Somers~\shortcite{somers2005round} and van Zaanen and Zwarts~\shortcite{van2006unsupervised} claimed that {\rtt} is inappropriate as a QE metric for translations.
The idea was supported by the low correlations between a {\bleu} score for the input and {\rtt} (\rttbleu) and a {\bleu} score for the reference and output.
However, {\bleu} is not an adequate metric to validate {\rtt} for QE.
When Aiken and Park~\shortcite{aiken2010efficacy} re-assessed {\rtt} with human judgments, there was a significant positive correlation between the human scores of round-trip translations and one-way translations.

Recently, {\rtt} has been employed for other purposes: generating paraphrased sentences and modeling purposes.
Yu et al.~\shortcite{yu2010confidence} exploit {\rtt}-based features to estimate the quality of spoken language translation and improve the accuracy of QE model.
Mallinson et al.~\shortcite{mallinson2017paraphrasing} reassess using {\rtt} for generating paraphrases in the context of NMT.
Junczys-Dowmunt and Grundkiewicz~\shortcite{junczys2017exploration} and Lichtarge et al.~\shortcite{lichtarge2019corpora} generate large amounts of artificial data to train an automatic post editing model and grammatical error correction, respectively.
Vaibhav et al.~\shortcite{vaibhav2019improving} also uses {\rtt} to augment bilingual data for NMT.
Lample et al.~\shortcite{lample2018phrase} measures {\rttbleu} for model selection purposes and Hassan et al.~\shortcite{hassan2018achieving} uses {\rtt} as a feature to re-rank translation hypotheses.

\subsection{Sentence Similarity Methods}

Lexical metrics, such as {\bleu}~\cite{papineni2002bleu} and {\chrf}~\cite{popovic2015chrf}, have long and widely been used for translation evaluation.
Both metrics compute strict matching between translation output and reference at the surface level.
{\bleu} counts the n-gram matches of the output and reference over the number of tokens of output as well as the length similarity of the output and reference. 
{\chrf} computes F-score based on character-level n-grams.
However, they cannot capture the semantic similarity of output and reference sentences beyond lexical relatedness or overlap.
In this sense, lexical-based metrics may not be the best way to measure the similarity of paraphrases.

{\bert}~\cite{devlin2019bert}, a pre-trained language representation model, made breakthroughs on many natural language processing tasks, including the sentence similarity prediction task~\cite{cer2017semeval}.
The methods using {\bert}'s embedding vectors were also introduced to MT evaluation, the task that needs semantic-level similarity measurement, and show the best performance~\cite{ma2019results,lo2019yisi,zhang2019bertscore}.
{\bertscore}~\cite{zhang2019bertscore} leverages {\bert} wordpiece embeddings to compute sentence similarity of two monolingual sentences.
When {\bertscore} is applied to the output and reference, it outperforms {\bleu} and {\chrf}.
Meanwhile, {\scshape sentence-Bert} ({\sbert})~\cite{reimers2019sentence}, a fine-tuned {\bert}, is introduced to derive more semantically meaningful sentence-level representation than {\bert}.
From the encouraging results of the embedding-based methods, we would expect the embeddings to catch the semantic similarity of input and round-trip sentences.


\section{{\rtt}-based QE Metrics}
\label{proposed-methods}

Given an input sentence $x=(x_1, x_2, x_3, ..., x_n)$ and a round-trip sentence $\hat{x}=(\hat{x}_1, \hat{x}_2, \hat{x}_3, ..., \hat{x}_m)$, an {\rtt}-based QE metric $f$ is a scalar function computing the similarity of $x$ and $\hat{x}$. 
We consider the scalar output as a quality for the translation of $x$.
The validity of $f$ is assessed primarily by Pearson correlation against the human judgments.

Previously, only surface-level similarity metrics were used for $f$. 
In this paper, we propose to use semantic-level metrics which can capture higher-level concepts of the similarity.
Detailed implementations are described in the Appendix~\ref{appendix:implementation}.


\subsection{Surface-level Metrics}

\paragraph{{\scshape RTT-Bleu} / {\scshape RTT-sentBleu}}
{\bleu}~\cite{papineni2002bleu} has originally designed to measure system-level translation performance.
To evaluate a sentence-level translation, {\sentbleu}, the smoothed version of {\bleu}, has been used~\cite{ma2019results,ma2018results}.
Since system-level {\bleu} and sentence-level {\bleu} exploit different computation method, we also separate {\bleu}-based {\rtt} QE metric for the system-level and sentence-level.
Specifically, {\rttbleu} is either {\bleu} or {\scshape sacreBleu-Bleu}~\cite{post2018call} on system-level input sentences and round-trip sentences while {\scshape RTT-sentBleu} is {\sentbleu} on a single input sentence and round-trip sentence.



\paragraph{\rttchrf} 
Sentence-level score is produced by {\chrf} and system-level score is the average of the segment score obtained by {\scshape sacreBleu-chrF\footnote{It is widely known that their scores are slightly different from the average of {\scshape chrF} even with the same parameters. 
Since {\scshape sacreBleu} is standard, we take {\scshape sacreBleu-chrF} for the system-level score.}}~\cite{post2018call}.


\subsection{Semantic-level Metrics}

In our settings, the semantic-level metrics are represented by the cosine similarity of {\sbert} embeddings and {\bertscore}. 
For all metrics, system-level score is an averaged sentence-level scores.



\paragraph{\rttsbert}
{\rttsbert} calculates the cosine similarity of $x$ and $\hat{x}$ embedding vectors extracted from {\sbert}~\cite{reimers2019sentence}.
We use a publicly available pre-trained {\scshape SBert}\footnote{\url{https://github.com/UKPLab/sentence-transformers}}.
Note that released models support Arabic, Chinese, Dutch, English, French, German, Italian, Korean, Polish, Portuguese, Russian, Spanish, and Turkish.


\paragraph{\rttbertscore}

{\rttbertscore} computes F-score based on wordpiece-level embedding similarities of $x$ and $\hat{x}$ weighted by inverse document frequency (idf), where each embedding is taken from {\bert}.
The idf weights penalize common wordpiece similarities, such as end of sentence symbols.
Given $L$ input sentences $\{x^k\}_{k=1}^{L}$, the idf score of $x_i$ is defined as:
\begin{equation*}
    \text{idf}(x_i) = - \log\frac{1}{L} \sum_{i=1}^{L} \mathbbm{1} [ x_i \in x^k ]
\end{equation*}

\section{Experimental Settings}

We compare {\rtt}-based semantic-level QE metrics to lexical-level QE metrics in various conditions.
Initially, we prepare different BT systems to see the impact of the BT system to the performance change in {\rtt}-based metrics. 
Then, with a suitable BT system, we observe the proposed metrics on WMT 2019 metrics task evaluation dataset.
We also examine whether our methods are biased to the certain type of FT system.
Furthermore, we investigate relations of the performance of {\rtt}-based QE metrics and their paraphrase detection ability.

\subsection{Data}

\paragraph{WMT metrics task evaluation set} 
The WMT19 dataset includes translations from English to Czech, German, Finnish, Gujarati, Kazakh, Lithuanian, Russian, and Chinese, and from the same set except Czech to English.
Translation outputs were provided by the WMT19 submitted systems where all were NMT.
Each system was not necessarily present in all language pairs, therefore, English--German received 22 submissions whereas German--English received 16 (see n in  Table~\ref{tab:system-level--mBERT-RTTBERT-RTTBLEU-RTTchrF}).
The human scores were gathered by using Direct Assessment (DA) for the translations of all systems on a scale of 0-100 points then standardized for each annotator. 
System's performance is an average over all assessed sentences produced by the given system and sentence-level golden truth is a relative ranking of DA judgments ({\scshape daRR}). 
In WMT19, QE-as-a-metrics were also assessed by the same standard as the reference-based metrics, namely Pearson correlation coefficient and Kendall’s $\tau$-like formulation against {\scshape daRR}, therefore, performance could be compared directly with {\bleu}~\cite{papineni2002bleu} and {\chrf}~\cite{popovic2015chrf}.

We also use WMT12 metrics task evaluation set to assess {\rtt}-based metrics on SMT. 
It includes translations from English to Czech, French, German, and Spanish and vice versa. 
We select English--German, German--English, and English--Czech which also appeared in WMT 2019.
The annotators were asked to evaluate sentences by ranking translated outputs from randomly selected 5 systems.
A ratio of wins is used for the system's performance~\cite{chris2012findings}.

\paragraph{PAWS} 
PAWS (Paraphrase Adversaries from Word Scrambling)~\cite{paws2019naacl} is a paraphrase identiﬁcation dataset constructed from sentences in Wikipedia (Wiki) and Quora Question Pairs (QQP) corpus.
We denote dataset as $\text{PAWS}_{\text{Wiki}}$ and $\text{PAWS}_{\text{QQP}}$ respectively.

Paraphrased sentences are generated by controlled word swapping and back translation, followed by fluency and paraphrase judgments by human raters.
Paraphrase and non-paraphrase pairs are mixed and to make dataset more challenging, both pairs have high lexical overlap.


\begin{table*}[t!]
\centering
\begin{adjustbox}{width=1.9\columnwidth, center}
\begin{tabular}{@{}lc|c|c|c|c|c|c@{}}
\toprule
\multicolumn{2}{c|}{Backward translations}          & \multicolumn{4}{c|}{Pearson correlations}      & \multicolumn{2}{c}{Variance ($\times 10^{-4}$)}                                                                                                                    \\ \midrule
\multicolumn{1}{l|}{Systems}               & \bleu  & \multicolumn{1}{l|}{\rttbleu} & \multicolumn{1}{l|}{\rttchrf} & \multicolumn{1}{l|}{\rttsbert} & \multicolumn{1}{l|}{\rttbertscore} & \multicolumn{1}{l|}{\rttsbert} & \multicolumn{1}{l}{\rttbertscore} \\ \midrule
\multicolumn{1}{l|}{Google}                & \textbf{46.96} & 0.797                         & 0.853                                                 & 0.941                          & 0.951                 & 5.08         & 1.96    \\
\multicolumn{1}{l|}{Microsoft}             & 42.68 & \textbf{0.845}                         & \textbf{0.877}                                            & \textbf{0.948}                          & 0.955         & 5.12         & 2.07                     \\
\multicolumn{1}{l|}{Amazon}                & 40.89 & 0.776                         & 0.804                                         & 0.941                          & \textbf{0.956}       & 4.86         & 1.88                        \\ \midrule
\multicolumn{1}{l|}{Facebook-FAIR}         & 42.17 & 0.788                         & 0.865                                           & 0.940                          & 0.934               & 4.84   & 1.27              \\
\multicolumn{1}{l|}{Transformer Big (100k)} & 38.96 & 0.739                         & 0.818                                         & 0.939                          & 0.937             & 4.58 & 1.57                \\
\multicolumn{1}{l|}{Transformer Big (40k)}  & 36.38 & 0.707                         & 0.795                                         & 0.938                          & 0.935        & 4.22 & 1.36                     \\
\multicolumn{1}{l|}{Transformer Big (20k)}  & 34.75 & 0.617                         & 0.759                                        & 0.931                          & 0.860       & 3.97 & 1.15                      \\
\multicolumn{1}{l|}{Transformer Big (10k)}  & 31.30  & 0.509                         & 0.749                                     & 0.908                          & 0.789        & 3.17 & 0.91                     \\ \bottomrule
\end{tabular}
\end{adjustbox}
\caption{\label{tab:bt-corr}
Performance of {\rtt}-based QE metrics on 22 English--German FT systems with various German--English BT systems. The variance of the best metrics, {\rttsbert} and {\rttbertscore}, are described, additionally.
}
\end{table*}


\subsection{Backward Translation (BT)}\label{settings:bt-systems}

To estimate the quality of MT systems with {\rtt}, a BT system is required. 
The choice of the BT system seemingly has the potential to largely affect the performance of {\rtt}-based QE metrics, so we run experiments to verify the effect of BT system qualities.
We compare two types of models--the system trained solely on WMT19 news translation task training corpus and online system--with different performance in terms of {\bleu}.
The BT systems trained on the WMT19 news dataset could be considered adequate to evaluate the WMT19 submitted FT systems since both systems are trained on the same domain.
On the other hand, online systems could also be desirable, because the online systems are trained on a huge amount of corpus mixed with various domains and would outperform the trained models on WMT19 dataset.
If the online systems show more favorable results, then {\rtt}-based QE metrics can be more practical in terms of the easy access to a BT system on any language pair.

To examine the impact of the BT systems, we choose English--German, which is the most submitted language pair.
For trained BT systems, we use Facebook-FAIR\footnote{Submitted model is publicly available via PyTorch \\(\url{https://pytorch.org/hub/pytorch_fairseq_translation}).}, the best system in WMT19 on German--English, and the Transformer Big model~\cite{vaswani2017attention} saved at 10k, 20k, 40k, and 100k iterations during training on the WMT19 corpus. 
Details of the Transformers are described in Appendix~\ref{appendix:deen-transformer-big}.
We also try three online systems, namely Google, Microsoft, and Amazon, showing different {\bleu} on WMT19 German--English evaluation set.
Each system was requested on Oct 2019, Nov 2019, and Dec 2019.


\subsection{Forward Translation (FT)}

The metric might penalize or favor a certain type of models.
For instance, {\bleu} has been argued to penalize rule-based systems against statistical systems~\cite{hovy2007investigating}.

To investigate whether {\rtt}-based QE metrics penalize FT systems based on their architecture, we assess {\rtt}-based QE metrics on both NMT and SMT.
As the all models submitted to WMT19 are NMT~\cite {ma2019results}, and the models submitted to WMT12 are SMT or rule-based model~\cite {chris2012findings}, we denote the former as NMT and the latter as SMT.
We compare {\rtt}-based QE metrics' performance with Pearson correlation coefficient for the language pairs both appeared on WMT19 and WMT12, English--Czech, English--German, and German--English.


\begin{table*}[t!]
\begin{threeparttable}
\centering
\begin{adjustbox}{width=\columnwidth * 2,center}
\begin{tabular}{lccccccccccccccccc}
\toprule
\multicolumn{1}{l|}{src lang}      & de            & fi            & gu            & kk            & lt            & ru             & \multicolumn{1}{c|}{zh}            & \multicolumn{1}{c|}{\multirow{3}{*}{avg. (std.)}} & en            & en            & en            & en            & en            & en            & en            & \multicolumn{1}{c|}{en}            & \multirow{3}{*}{avg. (std.)} \\
\multicolumn{1}{l|}{tgt lang}      & en            & en            & en            & en            & en            & en             & \multicolumn{1}{c|}{en}            & \multicolumn{1}{c|}{}                      & cs            & de            & fi            & gu            & kk            & lt            & ru            & \multicolumn{1}{c|}{zh}            &                       \\ \cmidrule(r){1-8} \cmidrule(r){10-17}
\multicolumn{1}{l|}{n}             & 16            & 12            & 11            & 11            & 11            & 14             & \multicolumn{1}{c|}{15}            & \multicolumn{1}{c|}{}                      & 11            & 22            & 12            & 11            & 11            & 12            & 12            & \multicolumn{1}{c|}{12}            &                       \\ \midrule
\multicolumn{1}{l|}{\scshape Bleu\tnote{*}}          & .849          & .982          & .834          & .946          & .961          & .879           & \multicolumn{1}{c|}{.899}          & \multicolumn{1}{c|}{.907 (.057)}                  & .897          & .921          & .969          & .737          & .852          & .989          & .986          & \multicolumn{1}{c|}{.901}          & .907 (.084)                  \\
\multicolumn{1}{l|}{\scshape chrF\tnote{*}}          & .917          & .992          & .955          & .978          & .940          & .945           & \multicolumn{1}{c|}{.956}          & \multicolumn{1}{c|}{.955 (.025)}                  & .990          & .979          & .986          & .841          & .972          & .981          & .943          & \multicolumn{1}{c|}{.880}          & .947 (.056)                  \\
\multicolumn{1}{l|}{\scshape sacreBleu-Bleu\tnote{*}}          & .813          & .985          & .834          & .946          & .955          & .873           & \multicolumn{1}{c|}{.903}          & \multicolumn{1}{c|}{.901 (.065)}                  & .994          & .969          & .966          & .736          & .852          & .986          & .977          & \multicolumn{1}{c|}{.801}          & .910 (.100)                  \\
\multicolumn{1}{l|}{\scshape sacreBleu-chrF\tnote{*}}          & .910          & .990          & .952          & .969          & .935          & .919           & \multicolumn{1}{c|}{.955}          & \multicolumn{1}{c|}{.947 (.028)}                  & .983          & .976          & .980          & .841          & .967          & .966          & .985          & \multicolumn{1}{c|}{.796}          & .937 (.074)                  \\\midrule
QE as a Metric                     &               &               &               &               &               &                &                                    &                                            &               &               &               &               &               &               &               &                                    &                       \\ \midrule
\multicolumn{1}{l|}{Individual Best\tnote{*}}        & .850          & .930          & .566          & .324          & .487          & .808           & \multicolumn{1}{c|}{.947} & \multicolumn{1}{c|}{- (-)}                  & .871          & .936          & .907          & .314          & .339          & .810          & .919          & \multicolumn{1}{c|}{.118}          & - (-)                 \\ \midrule
\multicolumn{1}{l|}{YiSi-2\tnote{*}}        & \textbf{.796}          & .642          & .566          & .324          & .442          & .339           & \multicolumn{1}{c|}{\textbf{.940}} & \multicolumn{1}{c|}{.578 (.232)}                  & .324          & .924          & .696          & .314          & .339          & .055          & .766          & \multicolumn{1}{c|}{.097}          & .439 (.319)                 \\ \midrule
\multicolumn{1}{l|}{\scshape RTT-Bleu}      & .130          & \textbf{.827}          & .641          & .859          & .596          & .295           & \multicolumn{1}{c|}{.825}          & \multicolumn{1}{c|}{.596 (.284)}                  & -.625         & .797          & .417          & .608          & .930          & -.334         & .572          & \multicolumn{1}{c|}{-.599}         & .221 (.637)                 \\
\multicolumn{1}{l|}{\scshape RTT-chrF}      & .495          & .810          & \textbf{.778}          & .776          & .692          & .524           & \multicolumn{1}{c|}{.875}          & \multicolumn{1}{c|}{.707 (.146)}                  & -.408         & .842          & .487          & .586          & .423          & -.153         & .750          & \multicolumn{1}{c|}{-.310}         & .277 (.493)                  \\ \midrule
\multicolumn{1}{l|}{\scshape RTT-SBert}     & .761          & -          & -          & -          & -          & \textbf{.867}           & \multicolumn{1}{c|}{.889}          & \multicolumn{1}{c|}{.839 (.005)}                  & .470          & .941          & .804          & .710          & .950          & \textbf{.410} & .833          & \multicolumn{1}{c|}{\textbf{.256}} & \textbf{.672} (.261)         \\
\multicolumn{1}{l|}{\scshape RTT-BERTScore} & .654          & .819          & .729          & \textbf{.889}          & \textbf{.712} & .816           & \multicolumn{1}{c|}{.912}          & \multicolumn{1}{c|}{\textbf{.790} (.095)}                  & \textbf{.473} & \textbf{.951} & \textbf{.819} & \textbf{.737} & \textbf{.966} & .342          & \textbf{.869} & \multicolumn{1}{c|}{.071}          & .654 (.324)                  \\ \bottomrule
\end{tabular}
\end{adjustbox}
\end{threeparttable}
\caption{\label{tab:system-level--mBERT-RTTBERT-RTTBLEU-RTTchrF}
Pearson correlations of system-level metrics with human judgments on WMT19. The best correlations of QE-as-a-metric within the same language pair are highlighted in bold. \mbox{*} denotes that reported correlations are from WMT19 metrics task~\cite{ma2019results}.
}
\end{table*}


\begin{table*}[t!]
\begin{threeparttable}
\centering
\begin{adjustbox}{width=\columnwidth * 2,center}
\begin{tabular}{lccccccccccccccccc}
\toprule
\multicolumn{1}{l|}{src lang}      & de            & fi            & gu            & kk            & lt            & ru            & \multicolumn{1}{c|}{zh}            & \multicolumn{1}{c|}{\multirow{3}{*}{avg. (std.)}} & en            & en            & en            & en            & en            & en            & en            & \multicolumn{1}{c|}{en}            & \multirow{3}{*}{avg. (std.)} \\
\multicolumn{1}{l|}{tgt lang}      & en            & en            & en            & en            & en            & en            & \multicolumn{1}{c|}{en}            & \multicolumn{1}{c|}{}                      & cs            & de            & fi            & gu            & kk            & lt            & ru            & \multicolumn{1}{c|}{zh}            &                       \\ \cline{1-8} \cline{10-17}
\multicolumn{1}{l|}{n}             & 85k           & 38k           & 31k           & 27k           & 22k           & 46k           & \multicolumn{1}{c|}{31k}           & \multicolumn{1}{c|}{}                      & 27k           & 100k          & 32k           & 11k           & 18k           & 17k           & 24k           & \multicolumn{1}{c|}{19k}           &                       \\ \midrule
\multicolumn{1}{l|}{\scshape sentBleu\tnote{*}}      & .056          & .233          & .188          & .377          & .262          & .125          & \multicolumn{1}{c|}{.323}          & \multicolumn{1}{c|}{.223 (.111)}                  & .367          & .248          & .396          & .465          & .392          & .334          & .469          & \multicolumn{1}{c|}{.270}          & .368 (.081)                  \\
\multicolumn{1}{l|}{\scshape chrF\tnote{*}}          & .122          & .286          & .256          & .389          & .301          & .180          & \multicolumn{1}{c|}{.371}          & \multicolumn{1}{c|}{.272 (.096)}                  & .455          & .326          & .514          & .534          & .479          & .446          & .539          & \multicolumn{1}{c|}{.301}          & .449 (.091)                 \\ \midrule
QE as a Metric                     &               &               &               &               &               &               &                                    &                                            &               &               &               &               &               &               &               &                                    &                       \\ \midrule
\multicolumn{1}{l|}{Individual Best\tnote{*}}        & .022          & .211          & -.001          & .096          & .075          & .089           & \multicolumn{1}{c|}{.253} & \multicolumn{1}{c|}{- (-)}                  & .069          & .236          & .351          & .147          & .187          & .003          & .226          & \multicolumn{1}{c|}{.044}          & - (-)                 \\ \midrule
\multicolumn{1}{l|}{YiSi-2\tnote{*}}        & \textbf{.068}          & .126          & -.001         & .096          & .075          & \textbf{.053}          & \multicolumn{1}{c|}{.253}          & \multicolumn{1}{c|}{.096 (.080)}                  & \textbf{.069} & \textbf{.212} & .239          & .147          & .187          & .003          & -.155         & \multicolumn{1}{c|}{.044}          & .093 (.131)                  \\ \midrule
\multicolumn{1}{l|}{\scshape RTT-sentBleu}  & -.169         & .095          & .111          & .140          & .086          & -.104         & \multicolumn{1}{c|}{.168}         & \multicolumn{1}{c|}{.047 (.130)}                 & -.122         & -.001         & .088          & .374          & .399          & -.110         & .157          & \multicolumn{1}{c|}{-.106}         & .085 (.211)                 \\
\multicolumn{1}{l|}{\scshape RTT-chrF}      & -.114         & .141          & \textbf{.184} & .130          & .099          & -.050         & \multicolumn{1}{c|}{.195}          & \multicolumn{1}{c|}{.083 (.119)}                  & -.093         & .055          & .119          & .395          & .310          & -.069         & .195          & \multicolumn{1}{c|}{-.075}         & .105 (.185)                  \\ \midrule
\multicolumn{1}{l|}{\scshape RTT-SBert}     & -.066         & -          & -          & -          & -          & -.013         & \multicolumn{1}{c|}{.225}          & \multicolumn{1}{c|}{.049 (.024)}                  & .025          & .169          & .268          & .444          & .503          & \textbf{.070} & .371          & \multicolumn{1}{c|}{\textbf{.064}} & .239 (.185)                  \\
\multicolumn{1}{l|}{\scshape RTT-BERTScore} & -.085         & \textbf{.185}          & .167          & \textbf{.204} & \textbf{.118}          & -.020         & \multicolumn{1}{c|}{\textbf{.255}} & \multicolumn{1}{c|}{\textbf{.118} (.125)}         & .065          & .194          & \textbf{.292} & \textbf{.494} & \textbf{.579} & .069          & \textbf{.391} & \multicolumn{1}{c|}{.056}          & \textbf{.268} (.205)        \\
\bottomrule
\end{tabular}
\end{adjustbox}
\end{threeparttable}
\caption{\label{tab:segment-level--mBERT-RTTBERT-RTTBLEU-RTTchrF}
Kendall's $\tau$ formulation of segment-level metric scores with human judgments on WMT19. The best correlations of QE-as-a-metric within the same language pair are highlighted in bold. For some language pairs, QE metrics obtain negative correlations. \mbox{*} denotes that reported correlations are from WMT19 metrics task~\cite{ma2019results}.
}
\end{table*}


\section{Results}

\subsection{Sensitivity to Backward Translation}{\label{result:bt}}

Due to the nature of {\rtt}-based QE metrics, a BT system is needed. 
We use a variety of BT systems in terms of the training recipe and {\bleu} on WMT19 German--English testset and observe the performance of {\rtt}-based QE metrics evaluated by Pearson correlation ($r$) with human scores~\cite{ma2019results,ma2018results}.
Note that well-performing metrics achieve high correlation coefficient.

According to Table~\ref{tab:bt-corr}, {\rttbertscore} and {\rttsbert} not only outperform the other metrics but are also robust to the type and performance of the BT systems.
On the other hand, {\rttbleu} and {\rttchrf} are sensitive to the performance of the BT systems, and the correlations fall behind {\rttbertscore} and {\rttsbert}.
Since BT systems scoring low {\bleu} have less chance of having same word orders in {\rtt} as with input sentences, the performance of surface-form metrics, {\rttbleu} and {\rttchrf}, decrease more sharply than {\rttsbert} and {\rttbertscore}.

The best correlation of each metric is accomplished when the online system is used for the BT system.
Even though Microsoft and Facebook-FAIR exhibit a similar {\bleu} score, metrics are more successful when using the Microsoft system.
This can be explained by a variance of {\rtt}-based QE metrics score.
In average, the variance of {\rttsbert} and {\rttbertscore} using the online BT systems is higher than that of trained ones.
The trained BT systems might over-translate a fault translation output similar to the original input, e.g., Kim Jong Un -- Kim -- Kim Jong Un, that make QE metrics hard to distinguish good systems to the bad ones.

Surprisingly, the best BT system in terms of {\bleu} does not always guarantee the best {\rtt}-based QE metrics. 
Despite Google's highest {\bleu} score, the performance of {\rtt}-based QE metrics is lower than or similar to that of Microsoft.
This assures that {\bleu} is not the only feature that affect the performance of the {\rtt}-based QE metrics.


\subsection{Performance across Language Pairs}\label{result:wmt19}

Provided from the results in Section~\ref{result:bt}, we use one of the online systems to get {\rtt} for all language pairs in WMT19.
Specifically, we use Google Translate, because of its coverage of supported language pairs and its overall performance across all language pairs.


\begin{table*}[t!]
\begin{threeparttable}
\centering
\begin{adjustbox}{width=1.7\columnwidth,center}
\begin{tabular}{@{}c|c|c|c|c|c|c@{}}
\toprule
\multirow{2}{*}{Language Pairs}      & \multicolumn{1}{l|}{\multirow{2}{*}{Systems (n)}} & \multicolumn{5}{c}{Pearson correlations}                  \\ \cmidrule(l){3-7} 
                                     & \multicolumn{1}{l|}{}        & {\bleu}             & {\rttbleu} & {\rttchrf} & {\rttsbert} & {\rttbertscore} \\ \midrule
\multicolumn{1}{l|}{\multirow{2}{*}{English--Czech}}  & SMT (12)         &   0.615                         & 0.261    & 0.342       & 0.482     & \textbf{0.620}         \\
\multicolumn{1}{l|}{}                & NMT (11)                          & 0.897                & -0.625   & -0.408      & 0.470     & \textbf{0.473}         \\ \midrule
\multicolumn{1}{l|}{\multirow{2}{*}{English--German}} & SMT (12)            & 0.582                              & 0.523    & 0.553    & 0.742     & \textbf{0.765}         \\
\multicolumn{1}{l|}{}                & NMT (22)                      & 0.921                   & 0.797   & 0.842     & 0.941     & \textbf{0.951}         \\ \midrule
\multicolumn{1}{l|}{\multirow{2}{*}{German--English}} & SMT (13)              & 0.841                            & 0.530    & 0.374      & \textbf{0.712}         & 0.682         \\
\multicolumn{1}{l|}{}                & NMT (16)                       & 0.849                   & 0.130    & 0.495    & \textbf{0.761}         & 0.654         \\ \bottomrule
\end{tabular}
\end{adjustbox}
\end{threeparttable}
\caption{\label{tab:smt-nmt-metric}
Pearson correlations of {\bleu} and {\rtt}-based QE metrics where FT systems are SMT and NMT. We reveal the number of systems in parenthesis.
}
\end{table*}


\begin{figure}[t]
    \centering
    \begin{subfigure}[b]{0.95\columnwidth}
        \centering
        \includegraphics[width=\textwidth]{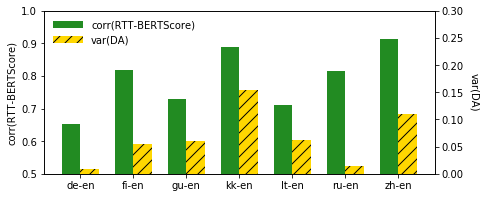}
        \includegraphics[width=\textwidth]{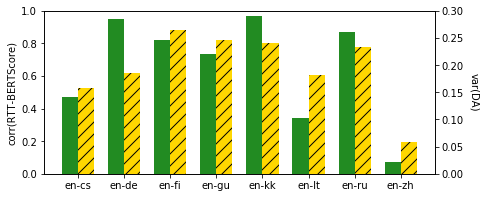}
    \end{subfigure}
    \caption{System-level correlations of {\scshape RTT-BERTScore} and variance of DA scores.}
    \label{fig:da-rtt-bert}
\end{figure}


\begin{figure}[t]
    \centering
    \includegraphics[width=0.9\columnwidth]{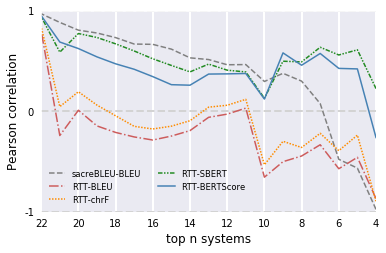}
    \caption{Pearson correlations of {\rtt}-based QE metrics and {\scshape sacreBLEU-BLEU} for English--German system-level evaluation for all systems (left) down to top 4 systems (right).}
    \label{fig:top-n-all}
\end{figure}


We conduct the same experiments as in the WMT19 metrics shared task to directly compare with the previous QE-as-a-metrics.
Individual best results of previous methods and {\yisi}-2 are provided to compare {\rtt}-based QE metrics within the same reference-free metrics.
Note that {\yisi} was the only QE-as-a-metric scoring on all language pairs, at the same time, achieving the best performance in total~\cite{ma2019results}.
We also include commonly used reference requiring metrics, namely {\bleu}, {\chrf}, {\scshape sacreBleu-Bleu}, and {\scshape sacreBleu-chrF}, to see how far QE metrics can get without reference translation.
The metrics are evaluated in system-level and sentence-level for all language pairs.
Specifically, Pearson correlation is applied to assess system-level metrics and Kendall's $\tau$-like formulation against the {\scshape daRR} to measure sentence-level metrics.

Table~\ref{tab:system-level--mBERT-RTTBERT-RTTBLEU-RTTchrF} illustrates the system-level correlations with human judgments on both to-English and out-of-English language pairs.
Across all language pairs except German--English, {\bert}-based {\rtt} QE metrics outperform {\rttbleu} and {\rttchrf}. 
For some language pairs, Gujarati--English, Kazakh--English, Russian--English, English--German, English--Gujarati, and English--Kazakh, {\rttbert}-based metrics show comparable result to {\bleu}, however, QE metrics still fall behind the reference-based metrics on average.
Surprisingly enough, the high Pearson correlation coefficients are mostly achieved on low-resource language pairs.
Results in Figure~\ref{fig:da-rtt-bert} suggest that this might due to the high variance of the system's DA scores which implies distinguishing good systems to the bad ones is relatively easy. 

To present a more reliable view, we draw plots of Pearson correlation while reducing MT systems to top $n$ ones as in Ma et al.~\shortcite{ma2019results}.
Figure~\ref{fig:top-n-all} depicts English--German, and all language pairs are in Appendix~\ref{appendix:top-n}. 
In general, correlations of {\scshape sacreBLEU-BLEU} and {\rtt}-based QE metrics tend towards 0 or negative, whereas the reference-based metric shows a rather continuous degradation than {\rtt}-based metrics. 
{\rttsbert} and {\rttbertscore} are better at retaining positive correlations compared to {\rttbleu} and {\rttchrf}, however, their consistency is weaker than {\scshape sacreBLEU-BLEU} except for some language pairs (English--German, English--Gujarati, English--Kazakh, and Finnish--English).
 
Metrics performance on sentence-level is described in Table~\ref{tab:segment-level--mBERT-RTTBERT-RTTBLEU-RTTchrF}\footnote{Instead of {\scshape Bleu}, we report {\scshape sentBleu}.}.
Sentence-level quality estimation is considered as a more difficult task than that of system-level. 
This is supported by the poor correlation coefficients of even {\scshape sentBleu} and {\chrf}.
Similar to the system-level results, QE metrics fall short of the reference-based metrics.
For language pairs with high DA score variance, again, {\rttbert}-based metrics provide comparable performance with reference-based metrics.


\begin{table*}[t!]
\centering
\begin{adjustbox}{width=2\columnwidth, center}
\begin{tabular}{@{}c|l|c|l|l@{}}
\toprule
\multicolumn{1}{l|}{Case} & Sentences      & \multicolumn{1}{l|}{Label} & \multicolumn{2}{c}{Ranks (out of 677)}                                                                                                             \\ \midrule
(a)                       & \begin{tabular}[c]{@{}l@{}}\textbf{sentence 1}: What are some example of \textit{deep web and dark web} ?\\\\ \textbf{sentence 2}: What are some example of \textit{dark web and deep web} ?\end{tabular}                & 1                          & \begin{tabular}[c]{@{}l@{}}{\sentbleu}: 534\\ {\chrf}: 416\end{tabular} & \begin{tabular}[c]{@{}l@{}}{\sbert}: 242\\ {\bertscore}: 101\end{tabular}  \\ \midrule
(b)                       & \begin{tabular}[c]{@{}l@{}}\textbf{sentence 1}: What was the CD that Deanna and family \textit{were} listening to \\ at the beginning of Try ( S5E15 ) of Season 5 of the Walking Dead and why was they listening to it ?\\\\ \textbf{sentence 2}: What was the CD that Deanna and family \textit{was} listening to \\ at the beginning of Try ( S5E15 ) of Season 5 of the Walking Dead and why were they listening to it ?\end{tabular} & 1                          & \begin{tabular}[c]{@{}l@{}}{\sentbleu}: 100\\ {\chrf}: 142\end{tabular} & \begin{tabular}[c]{@{}l@{}}{\sbert}: 14\\ {\bertscore}: 2\end{tabular}      \\ \midrule
(c)                       & \begin{tabular}[c]{@{}l@{}}\textbf{sentence 1}: How is dark/vacuum energy created with the universe conserved if it is not \textit{created} ? \\ Can infinite of these be \textit{conserved} ?\\\\ \textbf{sentence 2}: How is dark/vacuum energy created with the universe conserved if it is not \textit{conserved} ? \\ Can infinite of these be \textit{created} ?\end{tabular}          & 0                          & \begin{tabular}[c]{@{}l@{}}{\sentbleu}: 119\\ {\chrf}: 90\end{tabular}  & \begin{tabular}[c]{@{}l@{}}{\sbert}: 353\\ {\bertscore}: 501\end{tabular} \\ \bottomrule
\end{tabular}
\end{adjustbox}
\caption{\label{tab:qqp}
Example sentences on $\text{PAWS}_{\text{QQP}}$ dataset. Label 1 indicates paraphrased, and 0 represents dissimilarity. The higher the metric rank, the more similar the two sentences are.
}
\end{table*}

\begin{table*}[t!]
\centering
\begin{adjustbox}{width=2\columnwidth, center}
\begin{tabular}{@{}c|l|c|l|l@{}}
\toprule
\multicolumn{1}{l|}{Case} & Sentences                                                                                                                                                                                                                                                                                       & \multicolumn{1}{l|}{Label} & \multicolumn{2}{c}{Ranks (out of 8000)}                                                                                                                 \\ \midrule
(d)                       & \begin{tabular}[c]{@{}l@{}}\textbf{sentence 1}: Other famous spa towns include Sandanski , Hisarya , \textit{Kyustendil , Devin , Bankya} ,\\ Varshets , and Velingard .\\ \\ \textbf{sentence 2}: Other famous spa towns include Sandanski , Hisarya , \textit{Bankya , Devin , Kyustendil} ,\\ Varshets and Velingard .\end{tabular}  & 1                          & \begin{tabular}[c]{@{}l@{}}{\sentbleu}: 2510\\ {\chrf}: 1521\end{tabular} & \begin{tabular}[c]{@{}l@{}}{\sbert}: 884\\ {\bertscore}: 533\end{tabular}   \\ \midrule
(e)                       & \begin{tabular}[c]{@{}l@{}}\textbf{sentence 1}: \textit{Southport Tower} is the first new tower to be built at the \textit{southern} end of \textit{the Macleod Trail} \\ in almost 20 years .\\ \\ \textbf{sentence 2}: \textit{Macleod Trail} is the first new tower to be built at the \textit{south} end of \textit{Southport Tower} \\ in almost 20 years .\end{tabular} & 0                          & \begin{tabular}[c]{@{}l@{}}{\sentbleu}: 2942\\ {\chrf}: 1446\end{tabular} & \begin{tabular}[c]{@{}l@{}}{\sbert}: 4374\\ {\bertscore}: 7505\end{tabular} \\ \bottomrule
\end{tabular}
\end{adjustbox}
\caption{\label{tab:wiki}
Example sentences on $\text{PAWS}_{\text{WiKi}}$ dataset. Label 1 indicates paraphrased, and 0 represents dissimilarity. The higher the metric ranks, the more similar the two sentences are.
}
\end{table*}



\begin{table}[t!]
\centering\
\begin{adjustbox}{width=0.65\columnwidth,center}
\begin{tabular}{@{}l|cc@{}}
\toprule
\textbf{Metrics}      & \textbf{$\text{PAWS}_{\text{WiKi}}$} & \textbf{$\text{PAWS}_{\text{QQP}}$} \\ \midrule
\scshape sentBLEU      & 0.639          & 0.354         \\
\scshape chrF      & 0.584          & 0.405         \\\midrule
\scshape SBert     & 0.656          & \textbf{0.545}         \\ 
\scshape BERTScore & \textbf{0.718}          & 0.509         \\\bottomrule
\end{tabular}
\end{adjustbox}
\caption{\label{tab:paws-auc-pr}
AUC scores of precision-recall curves of {\bert}-based metrics on $\text{PAWS}_{\text{WiKi}}$ and $\text{PAWS}_{\text{QQP}}$ testing set. 
}
\end{table}


\subsection{Sensitivity to Forward Translation}

A certain type of FT system could be penalized by one metric according to its computation method. 
For this reason, we observe the performance of {\rtt}-based QE metrics on different FT systems: SMT and NMT.
SMT denotes the systems submitted to WMT12 and NMT represents the systems submitted to WMT19.
Same as the Section~\ref{result:wmt19}, we use Google Translate for the BT system and evaluate the metrics with Pearson correlation coefficient.
Results are shown in Table~\ref{tab:smt-nmt-metric}.

{\rttsbert} and {\rttbertscore} demonstrate the most promising performance regardless of the FT systems.
In contrast, {\rttbleu} and {\rttchrf} seem to favor SMT.
The correlation coefficient gap between {\bleu} and both {\rttbleu} and {\rttchrf} is smaller when FT system is SMT.


\subsection{Paraphrase Detection}

The results from all the previous sections consistently show the outstanding performance of {\rttsbert} and {\rttbertscore}.
We see this in a view of paraphrase detection ability of {\sbert} and {\bertscore}.
To confirm our assumption, we compare the area-under-curve (AUC) scores of precision-recall curves of the four metrics used to measure input and {\rtt} on PAWS dataset.
The higher the score is, the better the metric at paraphrase detection.
Table~\ref{tab:paws-auc-pr} depicts the results.
Note that {\sbert} indicates the cosine similarity of the embedding vectors of two sentence pairs extracted from the model.

As expected, {\bertscore} and {\sbert} outperform {\sentbleu} and {\chrf}.
In case (a) of Table~\ref{tab:qqp} and case (d) of Table~\ref{tab:wiki}, we can find {\sentbleu} and {\chrf} are sensitive to the change of word order.
Additionally, they are hard to distinguish paraphrases on long sentences.
From case (b), (c), (d), and (e), lexical-based metrics constantly view the sentences are not paraphrased.

The results imply that metrics capability to measure the semantic similarity is highly correlated to the performance of {\rtt}-based QE metrics.

\section{Conclusions}

We have presented round-trip translation for translation quality estimation. It can be used for QE with suitable semantic-level similarity metrics like {\sbert}~\cite{reimers2019sentence} and {\bertscore}~\cite{zhang2019bertscore}.
{\rttsbert} and {\rttbertscore} are robust to the choice of a BT system, which alleviates the disadvantages of {\rtt} being dependent on the BT system.
Moreover, both QE metrics significantly outperform the state-of-the-art QE metric, {\yisi}-2.
When the performance gap between the FT systems is large, {\rttsbert} and {\rttbertscore} provide comparable performance to {\bleu}.
They also perform well irrespective of the modeling architecture of FT systems.
In future work, it would be interesting to investigate when {\rtt}-based metrics become more reliable or unreliable. 

We find the high performance of {\rttsbert} and {\rttbertscore} is owing to {\sbert} and {\bertscore}'s ability to detect paraphrased sentences.
If better sentence similarity measurements appear, the performance of {\rtt}-based metrics would increase as well.
With the growing amount of the data and the advance of computing power, there certainly be a better measurement, thus {\rtt}-based QE metric is also promising.


\clearpage
\begin{appendices}

\section{Metrics Implementation}\label{appendix:implementation}

\paragraph{{\scshape RTT-Bleu} / {\scshape RTT-sentBleu}} 
{\scshape sacreBleu-Bleu}~\cite{post2018call} is used for system-level score and {\scshape sentBleu} for sentence-level which is a smoothed version of {\bleu}.
Following WMT19 metrics task~\cite{ma2019results}, we ran {\scshape sacreBleu-Bleu}\footnote{\url{https://github.com/mjpost/sacreBLEU}\label{fn:sacreBleu}} with \verb|BLEU+case.lc+| \verb|lang.de-en+numrefs.1+smooth.exp+| \verb|tok.intl+version.1.3.6| and {\scshape sentBleu} with \verb|sentence-Bleu| in the Moses toolkit\footnote{\url{https://github.com/moses-smt/mosesdecoder/tree/master/mert/sentence-bleu.cpp}}.
Since Chinese tokenization is not supported by the \verb|tok.intl| included in the package, we preprocess Chinese sentences with \verb|tokenizeChinese.py|\footnote{\url{http://hdl.handle.net/11346/WMT17-TVXH}}.

\paragraph{\scshape RTT-chrF} 
We also take the same computation procedure with WMT19 {\scshape sacreBleu-chrF}\textsuperscript{\ref{fn:sacreBleu}} and {\scshape chrF}\footnote{\url{https://github.com/m-popovic/chrF/chrF++.py}}.
We ran \verb|chrF3+case.mixed+lang.de-en+| \verb|numchars.6+numrefs.1+space.False+| 
\verb|tok.13a+version.1.3.6.| and python script \verb|chrF++.py| with the parameters \verb|-nw 0| \verb|-b 3| respectively.


\paragraph{\scshape RTT-SBert}
We use \verb|bert-large-nli-| \verb|mean-tokens| for English and \verb|distiluse-| \verb|base-multilingual| \verb|-cased| for the others.

\paragraph{\scshape RTT-BERTScore}
{\bertscore} is publicly available\footnote{\url{https://github.com/Tiiiger/bert_score}} and it uses the same model and layer applied in {\rttbert}. 

\section{German--English Transformer big model configurations}\label{appendix:deen-transformer-big}

Hyperparameters of German--English transformer model used in Section~\ref{settings:bt-systems} generally followed transformer big configuration of Vaswani et al.~\shortcite{vaswani2017attention}, except for three shared embedding matrices of encoder input, decoder input, and decoder output.
In other words, we set the matrices' variables independently.

For training data, we used all downloadable parallel corpus on WMT19 news translation task for German--English: Europarl, ParaCrawl, CommonCrawl corpus, News Commentary, Wiki titles, and Rapid corpus of EU press releases.
Then, we cleaned corpora by filtering sentence pairs whose token length ratio is bigger than 1.5 or less than 0.66 and left 37,066,883 parallel lines. 

We normalized corpora with \verb|normalize-| \verb|punctuation.perl| in the Moses toolkit\footnote{\url{https://github.com/moses-smt/mosesdecoder/blob/master/scripts/tokenizer/normalize-punctuation.perl}} and tokenized them using bype-pair encoding implemented in Google's SentencePiece\footnote{\url{https://github.com/google/sentencepiece}}. 
Encoding models for German and English are separately trained with vocabulary size 32K. 

Finally, we trained model with mini batch containing approximately 35K tokens of English and 35K of German for each iteration.

\onecolumn
\section{Correlations for Top-N Systems}
\label{appendix:top-n}
\subsection{de-en}
\begin{figure*}[h!]
    \centering
    \includegraphics[width=\textwidth]{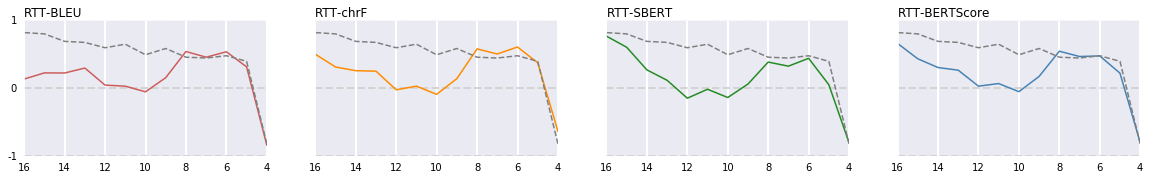}
\end{figure*}

\subsection{en-cs}
\begin{figure*}[h!]
    \centering
    \includegraphics[width=\textwidth]{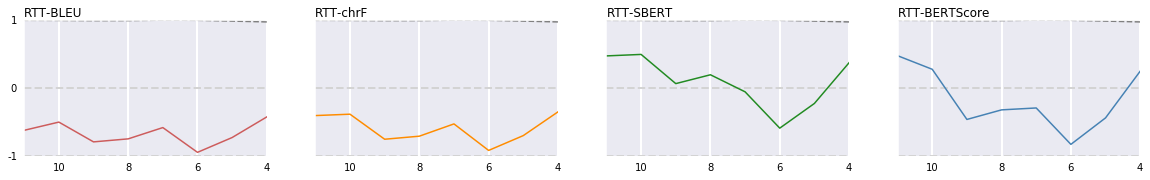}
\end{figure*}

\subsection{en-de}
\begin{figure*}[h!]
    \centering
    \includegraphics[width=\textwidth]{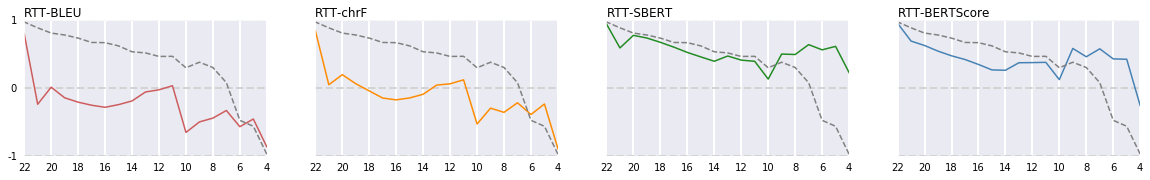}
\end{figure*}

\subsection{en-fi}
\begin{figure*}[h!]
    \centering
    \includegraphics[width=\textwidth]{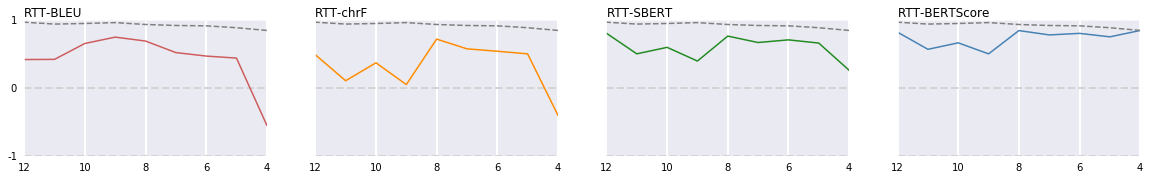}
\end{figure*}

\subsection{en-gu}
\begin{figure*}[h!]
    \centering
    \includegraphics[width=\textwidth]{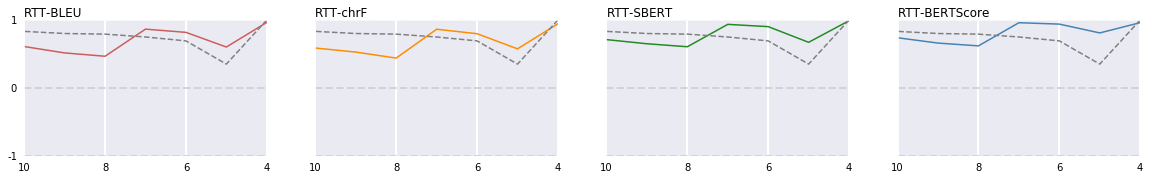}
\end{figure*}

\clearpage
\onecolumn
\subsection{en-kk}
\begin{figure*}[h!]
    \centering
    \includegraphics[width=\textwidth]{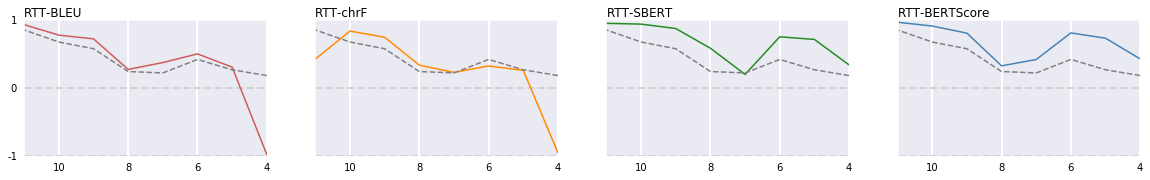}
\end{figure*}

\subsection{en-lt}
\begin{figure*}[h!]
    \centering
    \includegraphics[width=\textwidth]{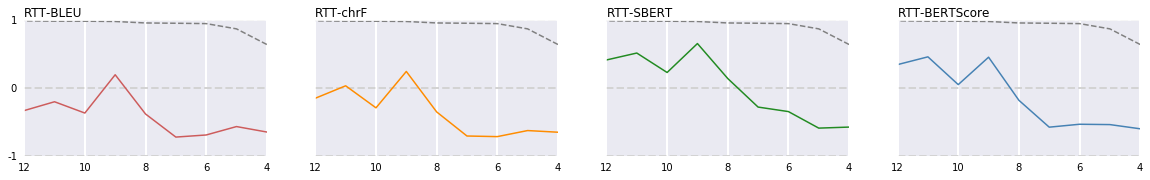}
\end{figure*}

\subsection{en-ru}
\begin{figure*}[h!]
    \centering
    \includegraphics[width=\textwidth]{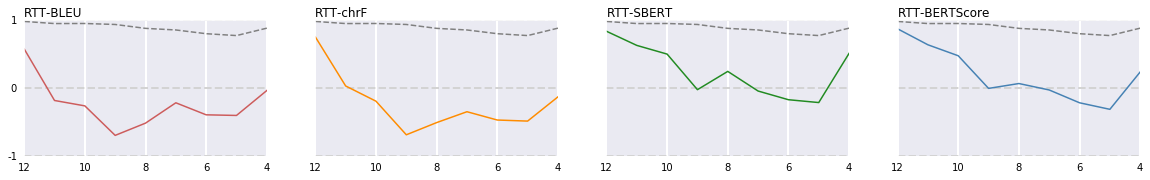}
\end{figure*}

\subsection{en-zh}
\begin{figure*}[h!]
    \centering
    \includegraphics[width=\textwidth]{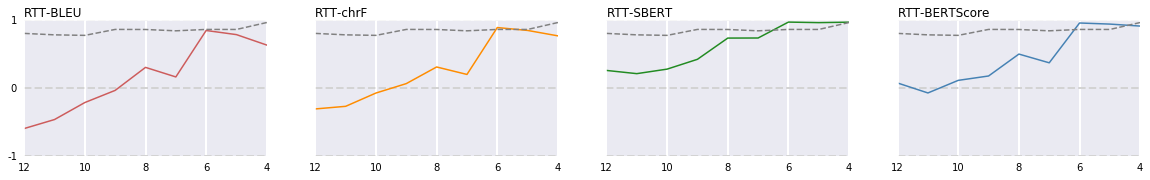}
\end{figure*}

\subsection{fi-en}
\begin{figure*}[h!]
    \includegraphics[width=0.75\textwidth,left]{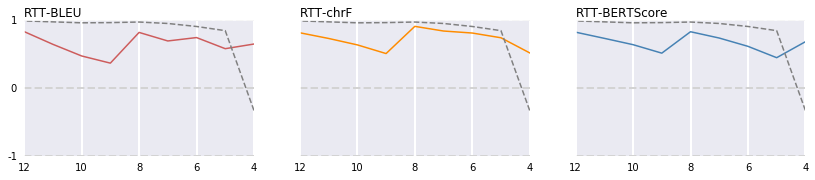}
\end{figure*}

\clearpage
\subsection{gu-en}
\begin{figure*}[h!]
    \includegraphics[width=0.75\textwidth,left]{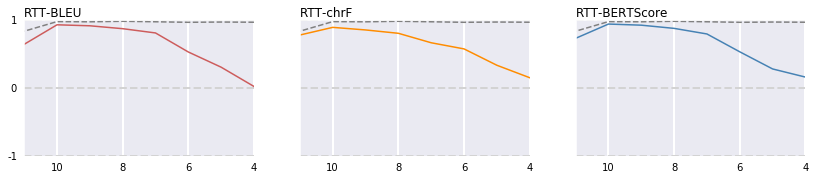}
\end{figure*}

\subsection{kk-en}
\begin{figure*}[h!]
    \includegraphics[width=0.75\textwidth,left]{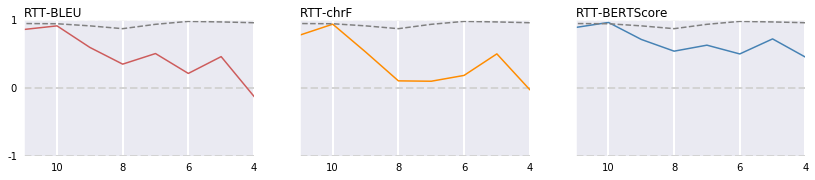}
\end{figure*}

\subsection{lt-en}
\begin{figure*}[h!]
    \includegraphics[width=0.75\textwidth,left]{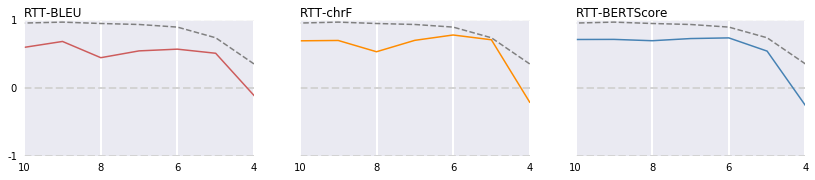}
\end{figure*}

\subsection{ru-en}
\begin{figure*}[h!]
    \centering
    \includegraphics[width=\textwidth]{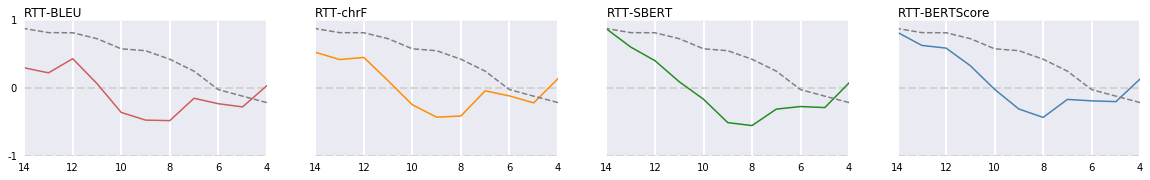}
\end{figure*}

\subsection{zh-en}
\begin{figure*}[h!]
    \centering
    \includegraphics[width=\textwidth]{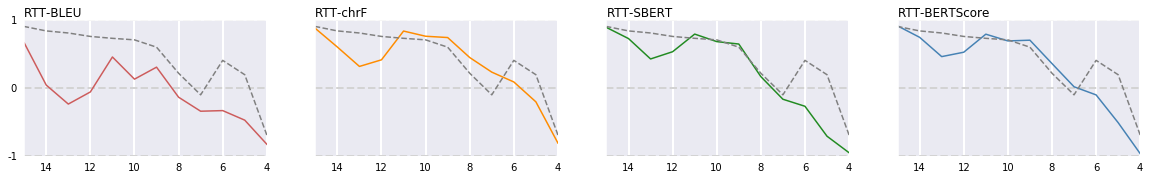}
\end{figure*}

\end{appendices}
\end{document}